\documentclass[letterpaper, 10 pt, conference]{ieeeconf}  

\IEEEoverridecommandlockouts                              

\usepackage{graphics} 
\usepackage{epsfig} 
\usepackage{mathptmx} 
\usepackage{times} 
\usepackage{hyperref}
\usepackage{amsmath}
\usepackage{amssymb}  

\newcommand{\gapbetweencaptionandtext}{-0.5cm}

\DeclareGraphicsExtensions{.pdf,.png,.jpg}

\usepackage{algorithm}
\usepackage{algpseudocode}
\algtext*{EndFor}

\usepackage{todonotes}
\presetkeys
    {todonotes}%
    {inline,figcolor=white}{}

\title{\LARGE \bf
Learning Dexterous Manipulation for a Soft Robotic Hand from Human Demonstrations
}

\author{Abhishek Gupta$^{1}$ \and Clemens Eppner$^{2}$ \and Sergey Levine$^{1}$ \and Pieter Abbeel$^{1}$
\thanks{$^{1}$Department of Electrical Engineering and Computer Sciences, University of California at Berkeley, CA, USA. 
This research was funded in part by ONR through a Young Investigator
Program award and by the Berkeley Vision and Learning Center (BVLC). {\tt\small \{abhigupta, svlevine, pabbeel\}@berkeley.edu}}%
\thanks{$^{2}$Robotics and Biology Laboratory, Technische Universit\"at Berlin, Germany. The author gratefully acknowledges financial support by the European Commission (SOMA, H2020-ICT-645599). \tt\small  clemens.eppner@tu-berlin.de}%
}

\begin{document}

\maketitle
\thispagestyle{empty}
\pagestyle{empty}

\begin{abstract}
Dexterous multi-fingered hands can accomplish fine manipulation behaviors that are infeasible with simple robotic grippers. However, sophisticated multi-fingered hands are often expensive and fragile. Low-cost soft hands offer an appealing alternative to more conventional devices, but present considerable challenges in sensing and actuation, making them difficult to apply to more complex manipulation tasks. In this paper, we describe an approach to learning from demonstration that can be used to train soft robotic hands to perform dexterous manipulation tasks. Our method uses object-centric demonstrations, where a human demonstrates the desired motion of manipulated objects with their own hands, and the robot autonomously learns to imitate these demonstrations using reinforcement learning. We propose a novel algorithm that allows us to blend and select a subset of the most feasible demonstrations, which we use with an extension of the guided policy search framework that learns generalizable neural network policies. We demonstrate our approach on the RBO Hand~2, with learned motor skills for turning a valve, manipulating an abacus, and grasping.
\end{abstract}

\section{Introduction}
Control of multi-fingered hands for fine manipulation skills is exceedingly difficult, due to the complex dynamics of the hand, the challenges of non-prehensile manipulation, and under-actuation. Furthermore, the mechanical design of multi-finger hands tends to be complex and delicate. Although a number of different hand designs have been proposed in the past~\cite{mouri2002anthropomorphic,shadowhand}, many of these hands are expensive and fragile.

\begin{figure}[t]
    \centering
    \includegraphics[width=.34\textwidth]{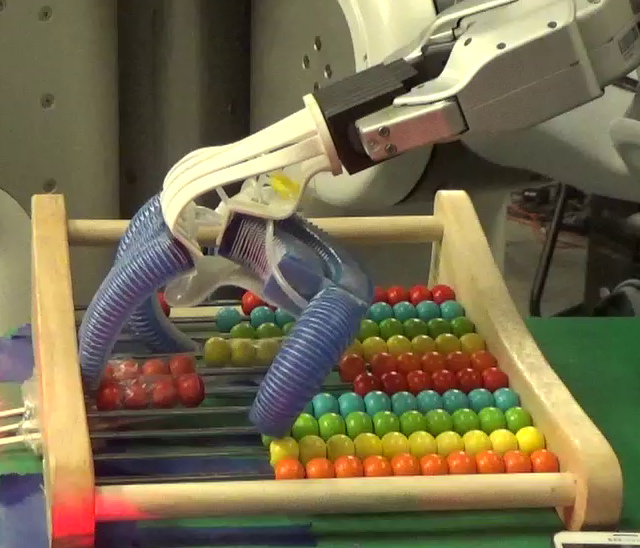}
    \caption{The RBO Hand~2 manipulating an abacus.}
    \label{fig:teaser}
    \vspace{\gapbetweencaptionandtext}
\end{figure}

In this work, we address the problem of autonomously learning dexterous manipulation skills with an inexpensive and highly compliant multi-fingered hand --- the RBO Hand~2~\cite{Deimel16-IJRR}. This hand (see Fig.~\ref{fig:teaser}) is actuated by inflating air-filled chambers. We cannot accurately control the kinematic finger motion, but can only actuate the hand by inflating or deflating the air chambers.
 Lack of sensing and precise actuation makes standard control methods difficult to apply directly to devices like this. Instead, we use an approach based on learning from demonstrations (LfD) and reinforcement learning (RL).

In LfD, the robot observes a human teacher solving a task and learns how to perform the demonstrated task and apply it to new situations. Demonstrations are typically given visually, by kinesthetic teaching, or through teleoperation. However, these techniques are difficult in case of the RBO Hand~2. A demonstrator cannot manually move all of the fingers of the hand for kinesthetic teaching, and the hand lacks position sensing to store such demonstrations.
Even direct teleoperation via intuitive interfaces, such as gloves or motion capture, is non-trivial, because although the RBO Hand~2 is anthropomorphic in design, its degrees of freedom do not match those of the human hand well enough to enable direct mapping of human hand motion.

However, the goal of most dexterous manipulation is to manipulate the poses of objects in the world, and a task can often be fully defined by demonstrating the motion of these objects. These demonstrations can be provided by putting trackers on the objects being manipulated and using a human demonstrator to physically move the objects along the desired trajectories. The object-centric demonstrations consist of just the trajectories of the object trackers, without any other states or actions.
These kinds of ``object-centric'' demonstrations are intuitive and easy to provide, but because the robot does not directly control the degrees of freedom of moving objects in the world, they cannot be imitated directly. Instead, we use reinforcement learning to construct controllers that reproduce object-centric demonstrations.

One crucial challenge that we must address to utilize object-centric demonstrations is to account for the mismatch between the morphology of the human expert and the robot. Since the robot cannot always reproduce all object-centric demonstrations, we propose a novel algorithm that automatically selects and blends those demonstrations that the robot can follow most closely, while ignoring irrelevant demonstrations that cannot be reproduced. 

Individual controllers may start from different initial conditions, such as different relative configurations of the hand and manipulated object, and thus may only be able to realize some subset of the demonstrated object-centric trajectories. Furthermore, some demonstrations may be difficult to achieve even from the same initial conditions as the ones they were demonstrated in, due to morphological differences between the robot and the human providing demonstrations, so the only way to determine if a demonstration is achievable is to attempt to imitate it. Our algorithm automatically determines which of the demonstrations are actually feasible to achieve from each initial state by alternating between reinforcement learning to learn the controllers and correspondence assignment to choose which demonstrations to follow from each state.

Our goal is to find a unified control policy that can generalize to a variety of initial states. To achieve generalization, we train a single nonlinear neural network policy to reproduce the behavior of multiple object-centric demonstrations. This approach follows the framework of guided policy search~(GPS~\cite{levine2014learning}), where multiple local controllers are unified into a single high-dimensional policy. This method is chosen because of its effectiveness in learning high dimensional control policies for real robots using a very small number of samples. However, unlike standard GPS, our approach requires only a set of object-centric demonstrations from a human expert to learn a new skill, rather than hand-specified cost functions.

The contributions of this paper are:
\begin{enumerate}
     \item{We propose a novel algorithm for learning from object-centric demonstrations. This algorithm enables complex dexterous manipulators to learn from multiple human demonstrations, selecting the most suitable demonstration to imitate for each initial state during training. The algorithm alternates between softly assigning demonstrations to individual controllers, and optimizing those controllers with an efficient trajectory-centric reinforcement learning algorithm.}
     \item{We demonstrate that a single generalizable policy can be learned from this collection of  controllers by extending the guided policy search algorithm to learning from demonstrations.}
     \item{We evaluate our approach by learning a variety of dexterous manipulation skills with the RBO Hand~2, showing that our method can effectively acquire complex behaviors for soft robots with limited sensing and challenging actuation mechanisms.}
\end{enumerate}

\section{Related Work}


\subsection{Dexterous Manipulation using Planning}

A variety of methods for generating manipulation behaviors with multi-fingered hands are based on planning. These approaches assume that a detailed model of the hand and object is available a priori. They generate open-loop trajectories that can be executed on real hardware. There exist planners that integrate contact kinematics, non-holonomic motion constraints, and grasp stability to come up with manipulation plans based on finger gaits~\cite{han1998dextrous}, rolling and sliding fingertip motions~\cite{cherif1999planning}, or nonprehensile actions involving only the palm~\cite{bai2014dexterous}. Optimization-based techniques~\cite{mordatch2012contact} have also been used for in-hand manipulation tasks. 
All of these approaches rely on detailed models, or make simplifying assumptions about the system. Modelling and simulating the behavior of a soft hand like the RBO Hand~2 is computationally expensive, since it requires finite-element method models~\cite{polygerinos2015fem} to achieve accuracy. Moreover, it is extremely hard to do accurate system identification on such systems.
In order to tackle this problem, our approach does not rely on detailed apriori models but learns the task-specific consequences of  actions from interactions of the real hardware with the environment, during a task.

\subsection{Reinforcement Learning for Manipulation}
In order to avoid planning with fixed handcrafted models, control policies that solve continuous manipulation problems can be found using reinforcement learning. A widely used approach is to learn the parameters of a dynamic motor primitive~\cite{ijspeert2002learning} (DMP) with relative entropy policy search~\cite{peters2010relative} or PI2~\cite{theodorouAISTATS2010}. This has been used to learn striking movements~\cite{mulling2013learning} and bi-manual transportation tasks~\cite{kroemer2015towards}. Although DMPs are often used to describe the kinematic state of a system, they can be used to generate compliant behavior for picking up small objects or opening doors~\cite{kalakrishnan2011learning}. However, DMP's typically require either a model of the system or the ability to control kinematic state, neither of which is straightforward on a soft hand that lacks position sensing.

Controllers for reaching and grasping have been learned by approximating the Q-function with a multilayer perceptron~\cite{lampe2013acquiring}. 
Policy search methods have succeeded in training neural network controllers to solve contact-rich peg-in-hole-like tasks~\cite{levine2015learning} based on positional or visual feedback~\cite{levine2015end}.

Some RL methods for manipulation have been applied to in-hand manipulation. Van Hoof et al.~\cite{vanHoof2015learning} learn a policy based on tactile feedback which lets an under-actuated hand slide cylindrical objects horizontally while being rolled between two fingers. Similar to our work is the learning method for an in-hand rotation tasks by Kumar et al.~\cite{kumaroptimal}. In contrast, we learn global policies that aim to generalize local solutions.

\subsection{Exploiting Human Demonstrations for Learning Manipulation Skills}

Learning from demonstrations has been effective in teaching robots to perform manipulation tasks with a limited amount of human supervision. By building statistical models of human demonstrations, gestures~\cite{calinon2007incremental} and dual-arm manipulations~\cite{asfour2008imitation} have been reproduced on robotic systems. Pure LfD can lead to suboptimal behavior when demonstrator and imitator do not share the same embodiment. To circumvent this problem the learning objective is often extended with additional feedback. This can be provided by a human, e.g. in the case of iteratively improving grasp adaptation~\cite{sauser2012iterative}. 
Alternatively, demonstrations can provide the coarse structure of a solution, while the details are iteratively refined and learned by the imitator itself. This has been shown for dexterous manipulation~\cite{prieur2012modeling} where an in-hand manipulation is broken down into a sequence of canonical grasps.
In combination with reinforcement learning, demonstrations often serve as an initial policy rollout or they constrain the search space by providing building blocks. This has been applied to  reaching motions~\cite{guenter2007reinforcement} and dynamic manipulation tasks.

\section{Algorithm Overview}

To find manipulation strategies for the RBO Hand~2 that solve different manipulation tasks, we take advantage of two general concepts: imitating human demonstrations and reinforcement learning. In order to learn from human demonstrations, we exploit task-specific information offered by human demonstrators using object-centric demonstrations, i.e. we only capture the motion of the object being manipulated, not hand-specific information. We use reinforcement learning to learn a policy which imitates these object centric demonstrations. However, due to kinematic and dynamic differences between the human hand and the RBO Hand~2, following some of these demonstrations might not be possible, and hence trying to imitate them closely is undesirable. We describe a novel demonstration selection algorithm that selects \textit{which} demonstration should be imitated, and use a reinforcement learning method to solve the problem of \textit{how} to imitate. 

We define our learning problem as optimizing a policy $\pi_{\theta}$ to perform the demonstrated task by learning from demonstrations. In order to learn this policy, we first train multiple different local controllers to imitate the most closely achievable demonstration from their respective initial states. This involves solving the joint problem of selecting the appropriate demonstration for each controller, and using reinforcement learning to train each controller to actually follow its chosen demonstration.
By modeling the objective as a minimization of KL divergence between a distribution of controllers and a mixture of demonstrations modeled as Gaussians, as shown in Section \ref{sec:LearningKLD}, this joint problem reduces to an alternating optimization between computing correspondence weights assigning a demonstration to each controller, and optimizing each controller using an optimal control algorithm. This algorithm can be used within the BADMM-based guided policy search framework~\cite{levine2015end}, to train a neural network policy $\pi_{\theta}$ to generalize over the learned controllers.
We propose a novel learning from demonstrations algorithm based on the GPS framework, which consists of three phases described in sections \ref{sec:LearningKLD}, \ref{sec:optimalcontrol}, and \ref{sec:suplearn}:
\begin{enumerate}
    \item Perform a weight assignment which computes soft correspondences between demonstrations and individual controllers~(Sec.~\ref{sec:LearningKLD}).
    \item With the soft correspondences fixed, solve an optimal control problem based on the correspondences and deviations from individual demonstrations~(Sec.~\ref{sec:optimalcontrol}).
    \item Perform supervised learning over the trajectory distributions from the optimal control phase, using the framework of BADMM-based GPS~(Sec.~\ref{sec:suplearn}).
\end{enumerate}

\setlength{\textfloatsep}{0.3cm}
\begin{algorithm}[!t]
{ \small
\caption{Guided policy search with demonstration selection
\label{alg:pseudocode}}
\begin{algorithmic}[1] 
\For{iteration $k = 1$ to $K$}
    \State Generate samples $\{\bar{\tau_j} \}$ from each controller $p_j(\bar{\tau})$  by running it on the soft hand.
    \State Compute soft correspondence weights $a_{ij}$ 
    \State Estimate system dynamics $p( x_{t+1} | x_t,  u_t)$ from $\{ \tau_j \}$
    \For{iteration $inner = 1$ to $n$}
        \State Perform optimal control to optimize objective defined in Section \ref{sec:LearningKLD}
        \State Perform supervised learning to match $\pi_{\theta}$ with the samples $\{ \bar{\tau_j}\}$ 
    \EndFor
\EndFor
\State \textbf{return} $\theta$\Comment{the optimized policy parameters}
\end{algorithmic}
}
\end{algorithm}

\section{Learning controllers from multiple demonstrations}
\label{sec:LearningKLD}

As the first step to generalizing dexterous manipulation skills, we learn a collection of controllers starting from different initial conditions, such that each controller imitates the demonstration which is most closely achievable from its initial condition. This problem can be cast as minimizing the
divergence between two distributions: one corresponding to the demonstrated trajectories, and one to the controllers. 

For our given dynamical system, we define the states to be~$x_t$, and the actions to be $u_t$ at every time step $t$. The system dynamics are specified by the model~$p(x_{t+1}|x_t, u_t)$.
Each controller $j$
is defined in terms of a conditional distribution $p_j(u_t|x_t)$, which along with the dynamics model $p(x_{t+1}|x_t, u_t)$ induces a distribution $p_j(\tau) = p_j(x_0) \prod p(x_{t+1}|x_t, u_t)p_j(u_t|x_t)$ over trajectories $\tau = {x_1, u_1, ..., x_T, u_T}$, where $T$ is the length of an episode. We define $p(\tau) = \sum_{j=1}^C \frac{1}{C} p_j(\tau)$ to be the uniform mixture of $C$ controllers $p_j(\tau)$.

Our state $x_t$ can be expressed as $x_t = [\bar{x_t},  x_t']$, where $\bar{x_t}$ denotes the "object-centric" parts tracking the manipulated objects and $x_t'$ is the rest of the state. In our experimental setting, $\bar{x_t}$ consists of positions and velocities of motion capture markers placed on manipulated objects.






As we are using object-centric demonstrations,
our objective is to match our controllers
with the demonstrations but only over the object centric elements~($\bar{x}$) of the state. For each controller $p(\tau)$, we can marginalize to obtain $p(\bar{\tau})$, which is a uniform mixture of $C$ distributions $p_j(\bar{\tau})$, such that $p(\bar{\tau}) = \sum_{j=1}^{C} w_j p_j(\bar{\tau})$ where $w_j = \frac{1}{C}$ and $\bar{\tau} = \{ \bar{x}_1, \bar{x}_2,...., \bar{x}_T\}$. This distribution is over just the object-centric trajectories $\bar{\tau}$.

 
The distribution of $D$ demonstrations over the trajectories~$\bar{\tau}$ is also modeled as a mixture, given by $d(\bar{\tau}) = \sum_{i}^{D} v_i d_i(\bar{\tau})$. Each $d_i(\bar{\tau})$ is defined as a multivariate Gaussian,
constructed according to $d_i(\bar{\tau}) = \mathcal{N}(\mu_i, \Sigma_i)$, where $\mu_i = \{\bar{x}_1, \bar{x}_2,..., \bar{x}_T\}$ is the trajectory of the objects recorded in each demonstration,
 and the covariance $\Sigma_i$ is a parameter that decides how closely the demonstration needs to be tracked by the controller.
The number of demonstrations and controllers do not have to be the same.

Our goal is to match the distribution of demonstrations with the distribution of controllers, which we formalize as a KL divergence objective: $\min_{p(\tau)} D_{KL} (p(\bar{\tau}) || d(\bar{\tau}))$. Although the objective is defined with respect to the object-centric distributions $p(\bar{\tau})$, the optimization is done with respect to the entire controller mixture $p(\tau)$ which includes other parts of the state, and actions.

Due to the mode seeking behavior of the KL divergence, this objective encourages each $p_j(\bar{\tau})$ to match the closest achievable demonstration. However, the KL divergence between mixtures cannot be evaluated analytically. Methods such as MCMC sampling can be used to estimate it, but we find a variational upper bound~\cite{hershey2007kldmixtures} to be the most suitable for our formulation. In order to simplify our objective, we decompose each mixture weight $w_j$ and $v_i$ into individual variational parameters $a_{ij}$ and $b_{ij}$, such that $\sum_{i} a_{ij} = w_j$ and $\sum_{j} b_{ij} = v_i$. We can rewrite
\begin{align*}
    D_{KL} \left(p(\bar{\tau}) || d(\bar{\tau})\right) &= 
    \int p(\bar{\tau}) \log{\frac{p(\bar{\tau})}{d(\bar{\tau})}} \\
    &=  \int -p(\bar{\tau}) \log{\frac{\sum_{i,j} b_{ij} d_i(\bar{\tau})}{p(\bar{\tau})}} \\
    &= - \int p(\bar{\tau}) \log{\sum_{i,j} \frac{b_{ij} d_i(\bar{\tau}) a_{ij} p_j(\bar{\tau})}{a_{ij} p_j(\bar{\tau}) p(\bar{\tau})}}.
\end{align*}
From Jensen's inequality we get an upper bound as follows:
\begin{align*}
    D_{KL} & \left(p(\bar{\tau}) || d(\bar{\tau})\right) \leq - \int p(\bar{\tau}) \sum_{i,j} \frac{a_{ij} p_j(\bar{\tau})}{p(\bar{\tau})} \log{\frac{b_{ij} d_i(\bar{\tau})}{a_{ij} p_j(\bar{\tau})}}\\
    &= - \sum_{i,j} \int p_j(\bar{\tau}) a_{ij} \log{\frac{b_{ij} d_i(\bar{\tau})}{a_{ij} p_j(\bar{\tau})}} \\
    &= \!\left[ \sum_{i,j} a_{ij} \int p_j(\bar{\tau}) \log{\frac{p_j(\bar{\tau})}{d_i(\bar{\tau})}} \right] - \!\left[ \sum_{i,j} a_{ij} \log{\frac{b_{ij}}{a_{ij}}} \right] \\
    &= \sum_{i,j} a_{ij} D_{KL}\left(p_j(\bar{\tau}) || d_i(\bar{\tau})\right) + D_{KL} \left(a || b\right)
\end{align*}
Thus, our optimization problem becomes
\begin{equation}
\label{eq:kdl_upper_bound}
\min_{p(\tau), a, b} \sum_{i,j} a_{ij} D_{KL}\left(p_j(\bar{\tau}) || d_i(\bar{\tau})\right) + D_{KL} \left(a || b\right).
\end{equation}
While the first term $\sum_{i,j} a_{ij} D_{KL}(p_j(\bar{\tau}) || d_i(\bar{\tau}))$ depends on the distribution $p(\tau)$, the second term $D_{KL} (a || b)$ depends on the mixing components $a_{ij}$ and $b_{ij}$ but is independent of the distribution $p(\tau)$.

We can perform the optimization in two alternating steps, where we first optimize $D_{KL} (p(\bar{\tau}) || d(\bar{\tau}))$ with respect to $a$, $b$, followed by an optimization of $D_{KL} (p(\bar{\tau}) || d(\bar{\tau}))$ with respect to $p(\tau)$, giving us a block coordinate descent algorithm in $\{a,b\}$ and $p$. The convergence of the algorithm is guaranteed by the convergence of a block coordinate descent method on a quasiconvex function and the fact that KL divergence is quasiconvex. The convergence properties of the weight assignment phase is shown in ~\cite{hershey2007kldmixtures}.

Intuitively, the first optimization with respect to $a, b$ is a weight assignment with the correspondence weight $a_{ij}$ representing the probability of assigning demonstration $i$ to controller $j$. The second optimization with respect to $p(\tau)$, keeps the correspondence parameters $a$, $b$ fixed, and finds optimal controllers using an optimal control algorithm to minimize a weighted objective specified in Eq. ~\ref{eq:meo}.

\subsubsection{Weight assignment phase}
The objective function $D_{KL} (p(\bar{\tau}) || d(\bar{\tau}))$ is convex in both $a$ and $b$, so we can optimize it by keeping one variable fixed while optimizing the other, and vice versa. We refer the reader to ~\cite{hershey2007kldmixtures} for further details on this optimization. This yields the following closed form solutions:
\[
b_{ij} = \frac{v_ia_{ij}}{\sum_{j'} a_{ij'}} \text{\quad and \quad}
a_{ij} = \frac{w_jb_{ij}e^{-D_{KL}(p_j(\bar{\tau})||d_i(\bar{\tau}))}}{\sum_{i'} b_{i'j}e^{-D_{KL}(p_j(\bar{\tau})||d_{i'}(\bar{\tau}))}}.
\]
In order to compute the optimal $a$ and $b$, we alternate between these updates for $a$ and $b$ until convergence.

\subsubsection{Controller optimization phase}
Once the optimal values for $a$ and $b$ have been computed, we fix these as correspondences between demonstrations and controllers and optimize~Eq.~\ref{eq:kdl_upper_bound} to recover the optimal $p(\tau)$.
As $a$ and $b$ are fixed, $D_{KL} (a || b)$ is independent of $p$. Hence, our optimization becomes:
\begin{align*}
    &\min_{p(\tau)} \sum_{i,j} a_{ij} D_{KL} (p_j(\bar{\tau}) || d_i(\bar{\tau})) \\
    &= \sum_{i,j} a_{ij} \left(-E_{p_j(\bar{\tau})} \left[\log{d_i(\bar{\tau})}\right] - H\left(p_j(\bar{\tau})\right) \right) \\
    &= \sum_j -w_j H(p_j(\bar{\tau})) - \sum_{i,j} a_{ij} E_{p_j(\bar{\tau})} \left[ \log{d_i(\bar{\tau})} \right].
\end{align*}
Factorizing the optimization to be independently over each of the controller distributions, for each controller  $p_j(\tau)$, we optimize the objective:
\begin{align}
    &- w_j H(p_j(\bar{\tau})) \!-\! \sum_i a_{ij} E_{p_j(\bar{\tau})} [\log{d_i(\bar{\tau})}] \\
    &\!=\! -w_j\left(H(p_j(\bar{\tau})) \!+\! \sum_i \frac {a_{ij}}{w_j} E_{p_j(\bar{\tau})} [\log{d_i(\bar{\tau})}]\right)
    \label{eq:meo}
\end{align}
In practice, the weight assignment is performed independently per time step, as the controllers we consider are time varying.

    
\section{Controller Optimization with an LfD objective}
\label{sec:optimalcontrol}
While the controller optimization phase could be performed with a variety of optimal control and reinforcement learning methods, we choose a simple trajectory-centric reinforcement learning algorithm that allows us to control systems with unknown dynamics, such as soft hands. Building on prior work, we learn time-varying linear Gaussian controllers by using iteratively refitted time-varying local linear models~\cite{levine2014learning}. This is predicated on the assumption that the system has Gaussian noise, which has been shown to work well in practice for robotic systems ~\cite{levine2014learning}.
The derivation follows prior work, but is presented here for the specific case of our LfD objective. Action-conditionals for the time-varying linear-Gaussian controllers are given by
\begin{equation*}
    p_j(u_t | x_t) = \mathcal{N} \left( K_{jt} x_t + k_{jt}, C_{jt} \right)
\end{equation*}
where $K_{jt}$ is a feedback term and $k_{jt}$ is an open loop term. Given this form, the maximum entropy objective~(Eq.~\ref{eq:meo}), can be optimized using differential dynamic programming~\cite{opac-b1078358,levine2013guided}. As $d_i(\bar{\tau})$ is a multivariate Gaussian $\mathcal{N}(\mu_i, \Sigma_i)$, we can rewrite the optimization problem in ~Eq.~\ref{eq:meo} as
\begin{equation*}
    \min_{p_j(\tau)} \sum_{t, i} \frac {a_{ijt}}{\sum_{i'}a_{i'jt}} E_{\bar{x_t} \sim p_j(\bar{\tau})} \left[\!\frac{1}{2}(\bar{x_t}\!-\! \mu_{it})^T\Sigma_i^{-1}(\bar{x_t}\!-\! \mu_{it})\!\right]\!-\!H(p_j(\bar{\tau}))
\end{equation*}
where we express the objective as a sum over individual time steps.
In this maximum entropy objective, the cost function defined as the expectation of a sum of $l_2$ distances of trajectory samples to each demonstration, weighted by the normalized correspondence weights $\frac {a_{ij}}{\sum_{i'} a_{i'j}}$. The trajectory samples denote the trajectories of the objects which we recover through object markers,
and we compute the $l_2$ distance of these samples
to the object-centric demonstrations.
Under linearized dynamics, this objective can be locally optimized using LQG \cite{li2004iterative}. However, for robots like the RBO Hand~2, the dynamics are complex and difficult to model analytically. Instead, we can fit a time-varying locally linear model of the dynamics, of the form $p(x_{t+1}|x_t,u_t) = \mathcal{N}(f_{xt}x_t + f_{ut}u_t|C_d)$, to samples obtained by running the physical system in the real world. The dynamics matrices $f_{xt}$ and $f_{ut}$ can then be used in place of the system linearization to optimize the controller objective using LQG~(for details see~\cite{li2004iterative,levine2014learning}). Important to note here is that the iLQG optimization learns~$K_{jt}$, $k_{jt}$ and $C_{jt}$ for the trajectory controller, while the term $\frac {a_{ij}}{\sum_{i'} a_{i'j}}$ is learned in the weight assignment phase and kept fixed for the iLQG optimization. 

One issue with optimizing a controller using fitted local dynamics models is that the model is only valid close to the previous controller. The new controller might visit very different states where the fitted dynamics are no longer valid, potentially causing the algorithm to diverge. To avoid this, we bound the maximum amount the controller changes between iterations. This can be expressed as an additional constraint on the optimization:
\begin{align}\label{eq:constklopt}
 D_{KL}\left(p_j\left(\tau\right)||\hat{p}_j\left(\tau\right)\right) < \epsilon,
\end{align}
where $\hat{p}_j(\tau)$ is the previous trajectory-controller and $p_j(\tau)$ the new one. As shown in \cite{levine2015end}, this constrained optimization problem can be formulated in the same maximum entropy form as Eq.~\ref{eq:meo}, using Lagrange multipliers, and solved via dual gradient descent~(for details and a full derivation see~\cite{levine2014learning,levine2015end}). In practice, each iteration of this controller optimization algorithm involves generating $N$ samples on the real physical system by running the previous controller, fitting a time-varying linear dynamics model to these samples as in previous work~\cite{levine2014learning}, and optimizing a new controller $p_j(\tau)$ by solving the constrained optimization using dual gradient descent, with LQG used to optimize with respect to the primal variables $K_{jt}$, $k_{jt}$, and $C_{jt}$. This can be viewed as an instance of model-based reinforcement learning.

\section{Supervised Learning using GPS}
\label{sec:suplearn}

The multiple controllers defined in the previous section learn to imitate the most closely imitable demonstration from their individual starting positions. However, given an unseen initial state, it is not clear which controller $p_j(\tau)$ should be used. For effective generalization, we need to obtain a single policy $\pi_{\theta}(u_t|x_t)$ that will succeed under various conditions. To do this, we extend the framework of GPS~\cite{levine2015end} to combine controllers into a single nonlinear neural network policy.

We learn the parameters $\theta$ of a neural network $\pi_{\theta}$ to match the behavior shown by the individual controllers by regressing from the state $x_t$ to the actions $u_t$ taken by the controllers at each of the $N$ samples generated on the physical system. Simply using supervised learning is not in general guaranteed to produce a policy with good long-horizon performance. In fact, supervised learning is effective only when the state distribution of $\pi_{\theta}$ matches that of the controllers $p_j(\tau)$. To ensure this, we use the BADMM-based variant of GPS~\cite{levine2015end}, which modifies the cost function for the controllers to include a KL-divergence term to penalize deviation of the controllers from the latest policy $\pi_{\theta}$ at each iteration. This is illustrated in Algorithm~(\ref{alg:pseudocode}), by first assigning correspondences between demonstrations and controllers, then alternating between trajectory optimization and supervised learning at every iteration, eventually leading to a good neural network policy~$\pi_{\theta}$. For further details on the guided policy search algorithm, we refer the reader to prior work~\cite{levine2015end}. 

\section{RBO Hand~2 and System Setup}

\begin{figure}[t]
    \centering
    \hfill
    \includegraphics[height=2.5cm]{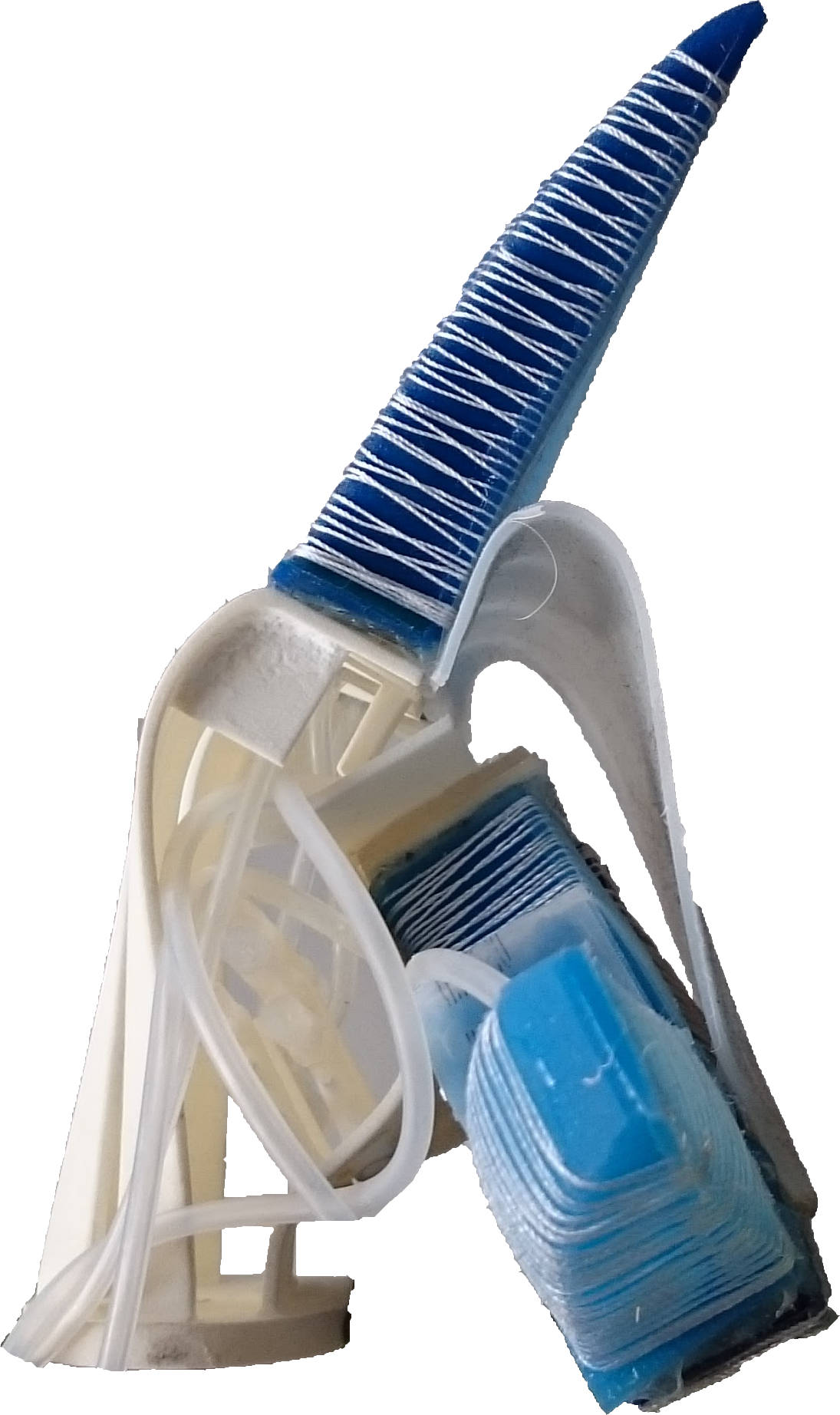}
    \hfill
    \includegraphics[height=2.5cm]{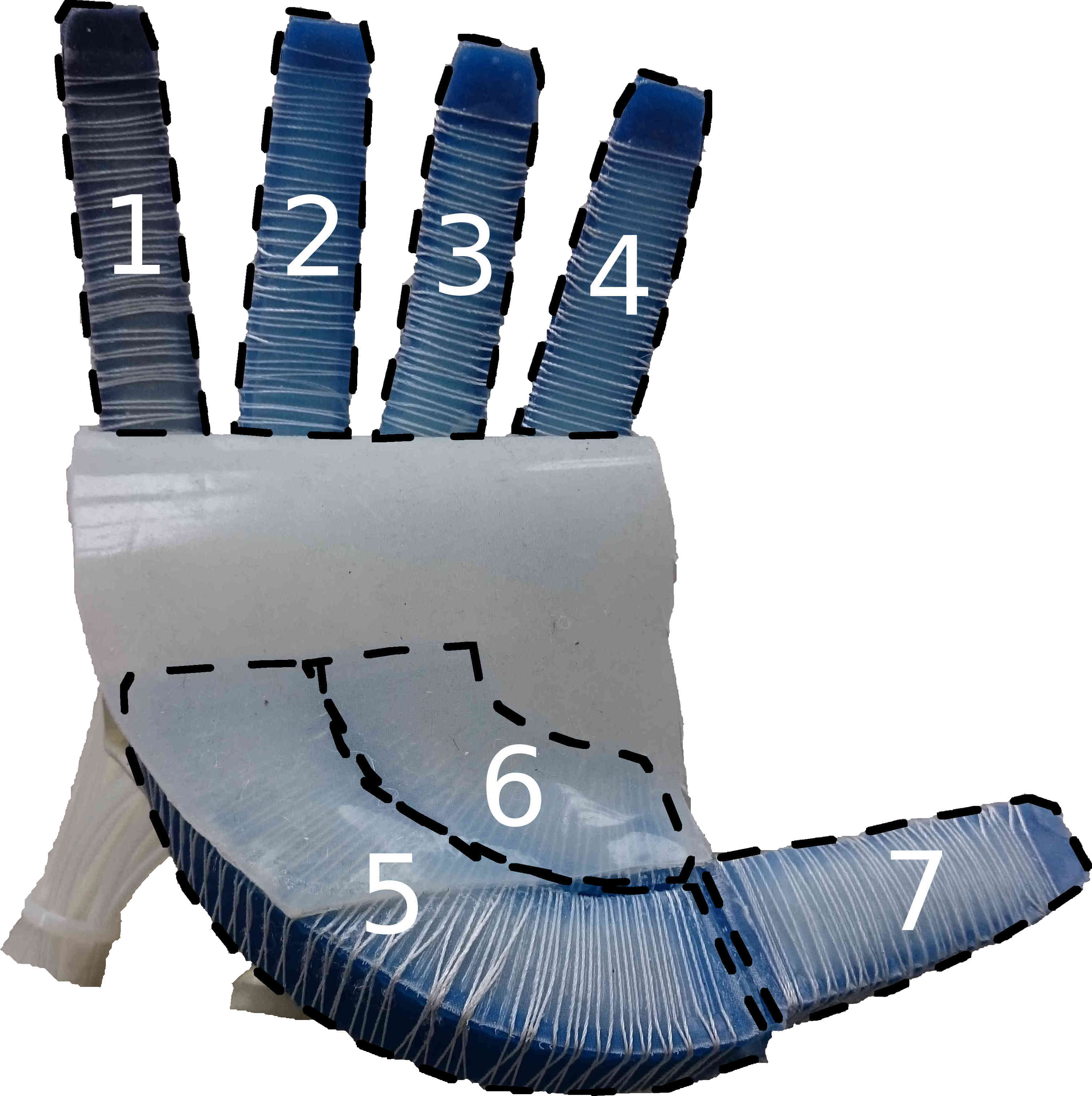}
    \hfill
    \includegraphics[height=2.5cm]{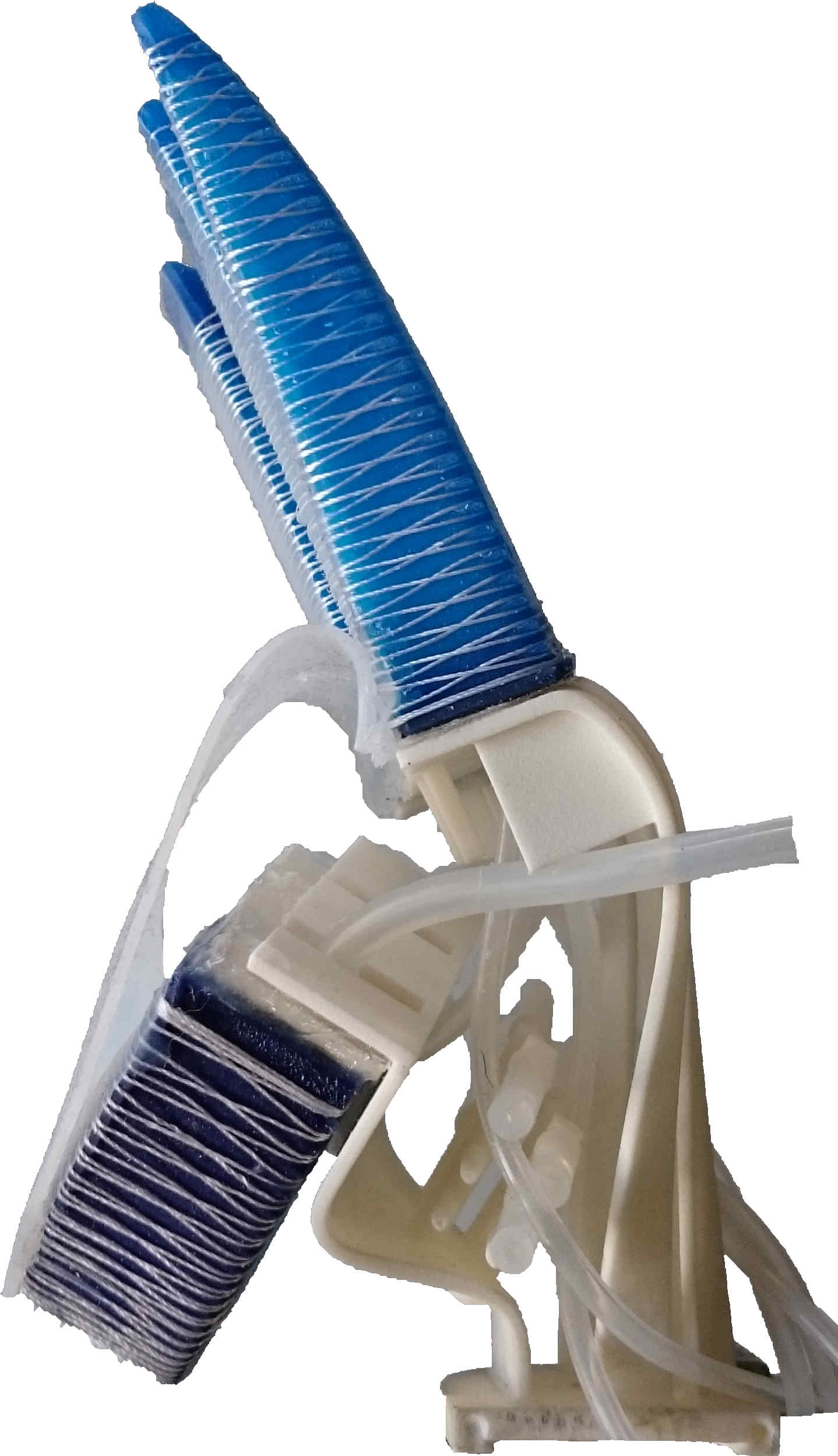}
    \hfill
    \caption{The RBO Hand~2 is an anthropomorphic pneumatically actuated soft hand consisting of seven actuators. Three of them form the palm and thumb. The air chambers can be physically coupled or actuated separately.}
    \label{fig:rbo_hand_2}
\end{figure}

\begin{figure*}[h]
    \centering
    \includegraphics[width=.3\textwidth]{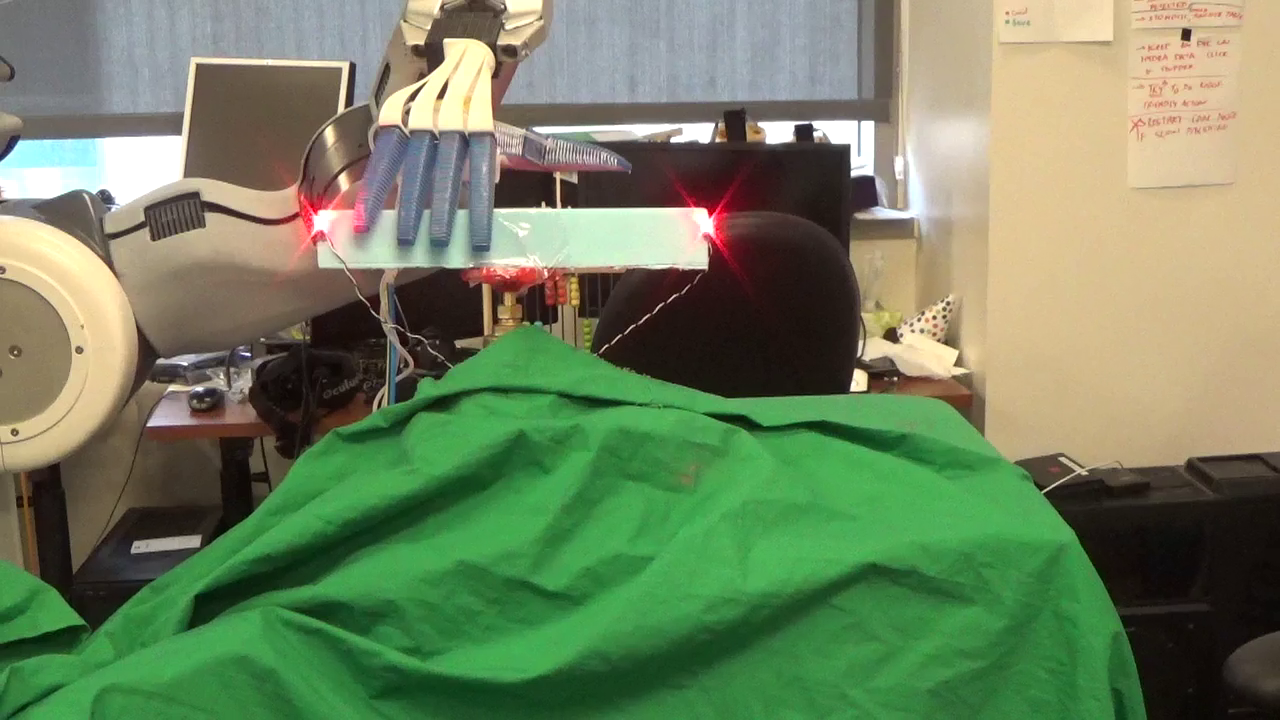}
    \hfill
    \includegraphics[width=.3\textwidth]{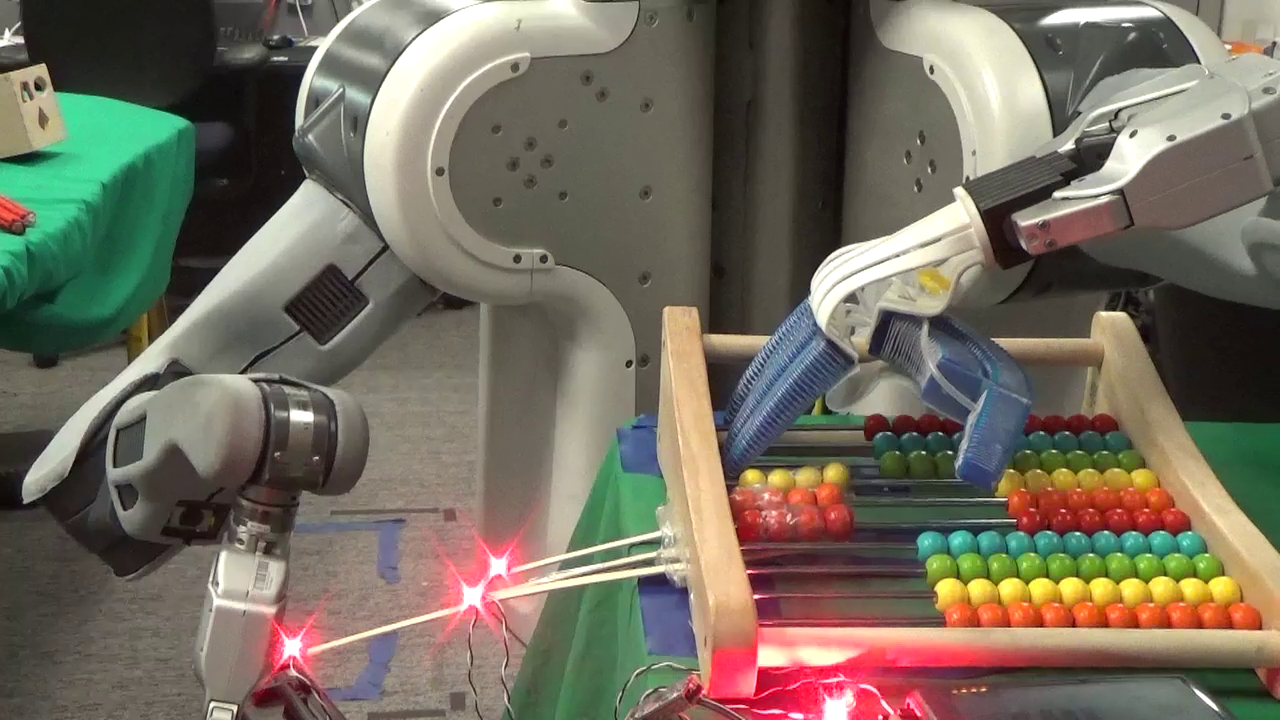}
    \hfill
    \includegraphics[width=.3\textwidth]{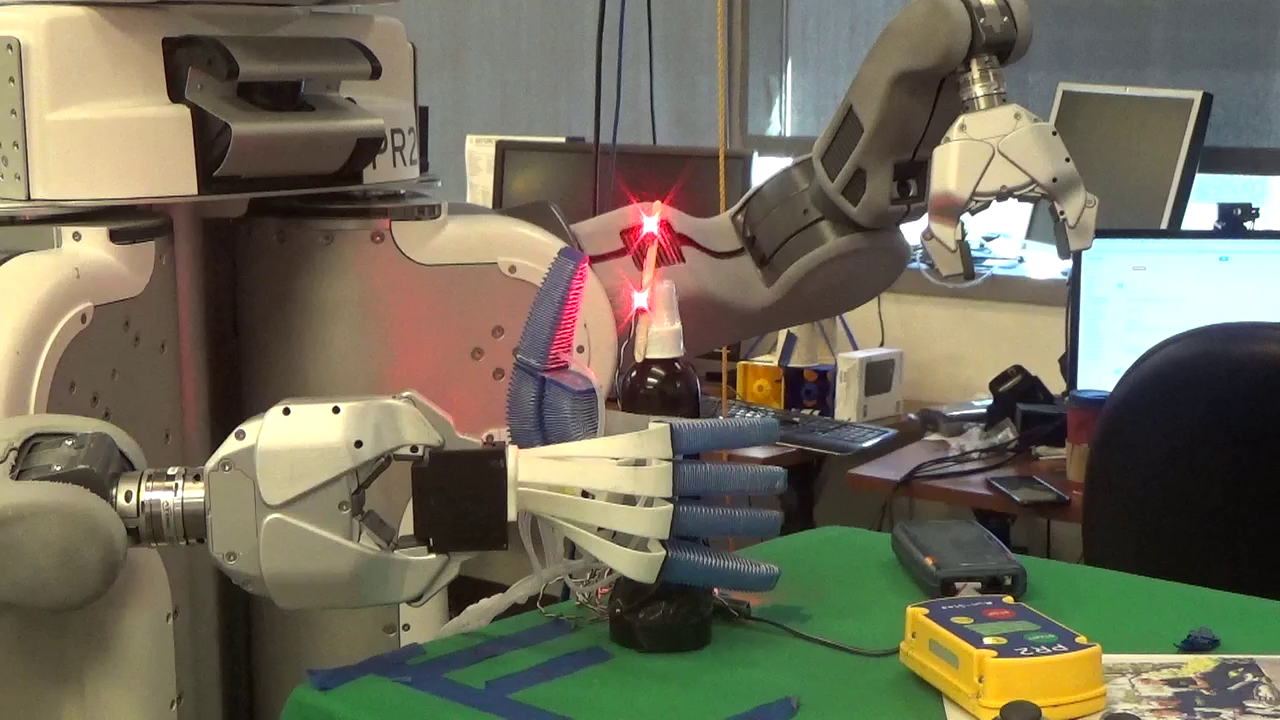}
    \caption{The three manipulation tasks used in our experiments: Turning a valve, pushing beads on an abacus, and grasping a bottle from a table.}
    \label{fig:experimentalsetups}
    \vspace{\gapbetweencaptionandtext}
\end{figure*}
The RBO Hand~2 is an inexpensive, compliant, under-actuated robotic hand which has been shown to be effective for a variety of grasping tasks~\cite{Deimel16-IJRR}.
The hand consists of a polyamide scaffold to which multiple pneumatic actuators are attached~(see~Fig.~\ref{fig:rbo_hand_2}).
Each of the four fingers is a single actuator, while the thumb consists of three independent pneumatic actuators.
This makes the thumb the most dexterous part of the hand, achieving seven out of eight configurations of the Kapandji test~\cite{kapandji_cotation_1986}. 
The actuators are controlled via external air valves and a separate air supply.
Control is challenging since the air valves can only be either fully closed or open and have a switching time of $\sim 0.02$s.
Each actuator has a pressure sensor located close to the air valve.

The hand is controlled by specifying valve opening durations to either inflate or deflate a single actuator.
We turn the discrete valve actions into a continuous control signal using pulse width modulation.
Given a constant frequency of 5Hz, the control signal is interpreted as the duration the inflation (if it is positive) or deflation (negative) valve is opened during a single time step. 
To ensure that the control signal does not exceed the duration of a single time step we apply a sigmoid function to the commands from the learning algorithm.

The positions and velocities of the manipulated objects are captured in real time with a PhaseSpace Impulse system, which relies on active LED markers. The state $x_t$ of our system is the concatenation of the seven pressure readings of the hand, their time derivatives, the 3D positions and velocities of markers attached to the object, and joint angles of the robot arm (depending on the task). We placed no LED markers on the hand itself, only the object was equipped to record object-centric demonstrations, and the positions and velocities of these markers constitute object-centric state $\bar{x_t}$.

\section{Experiments}
\noindent We evaluated our algorithm on a variety of manipulation and grasping tasks. Our experiments aim to show that 
\begin{enumerate}
    \item It is possible to perform fine manipulation with the RBO Hand~2.
    \item Our algorithm can learn feedback policies from demonstrations that perform nearly as well as an oracle with the correct demonstrations (depending on the context) manually assigned to controllers.
    \item A single neural network policy learned from demonstrations with our method is able to generalize to different initial states.
\end{enumerate}
We will evaluate our learning approach on three different tasks: turning a valve, pushing beads on an abacus, and grasping a bottle~(see Fig.~\ref{fig:experimentalsetups}). A video of the experiments can be found at \url{https://youtu.be/XyZFkJWu0Q0}. For the first two tasks, we compare our method to the following baselines:\\

\noindent\textbf{Hand designed baseline}: A controller with a hand-designed open loop policy. In the case of the abacus task, we evaluate the performance of two different strategies for simple hand-designed baselines.\\
\textbf{Single demo baseline}: A single controller trained to imitate a single demonstration. We use two separate baselines which are trained to follow different demonstrations.\\
\textbf{Oracle}: Depending on the context we manually assign the correct achievable demonstration to controllers. This comparison is useful to test whether the correspondence assignments are accurate. 

\subsection{Turning a valve}

\subsubsection{Experimental setup}
Rotating a gas valve is a challenging task since it involves coordinating multiple fingers. Our valve consists of a blue horizontal lever that increases its range of motion~(Fig.~\ref{fig:experimentalsetups}). Varying wrist positions along the lever require different finger motions to rotate it.

We mound the RBO Hand~2 on a PR2 robot arm, with the objective to rotate the valve away from the initial center position in either direction, using just its fingers. The arm is kept stationary for each episode, but changes positions between the training of different controllers. The joint angles are part of the state to determine the relative position of the hand with respect to the valve. 

A human demonstrated three different valve rotations with their own hand, while two LED markers tracked the motion of the lever. Two demonstrations were of the valve rotating clockwise and anti-clockwise at the same position, and a third demonstration with the valve placed at a different position and rotated anticlockwise. All three demonstrations are valid for the task, but not all of them are achievable from every training position. Our algorithm trained three individual controllers and a single neural network policy to generalize over them.


\begin{figure*}[!h]
    \centering
    \begin{tabular}{ c p{4.1cm} p{4.1cm} p{4.1cm} p{4.1cm} c }
     &
     \includegraphics[width=.17\textwidth]{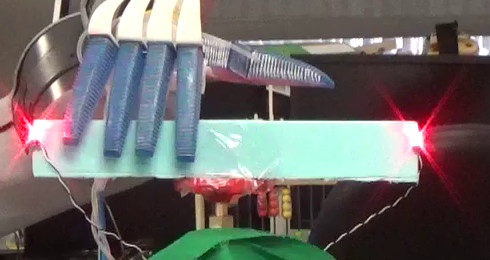} & \includegraphics[width=.17\textwidth]{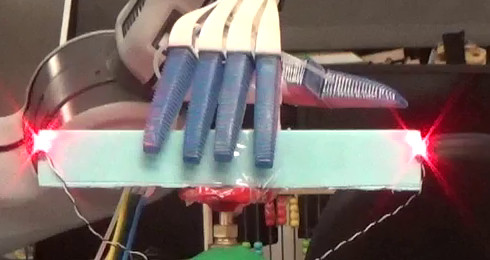} & \includegraphics[width=.17\textwidth]{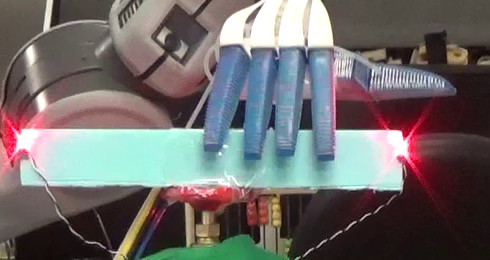} &
     \includegraphics[width=.17\textwidth]{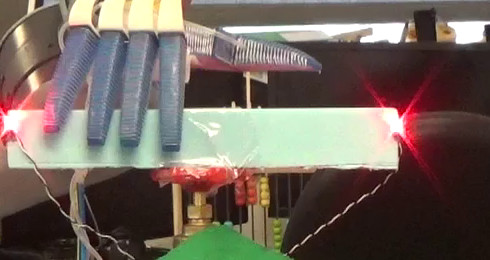} &
     
    \end{tabular}
    \\
    \includegraphics[width=.245\textwidth]{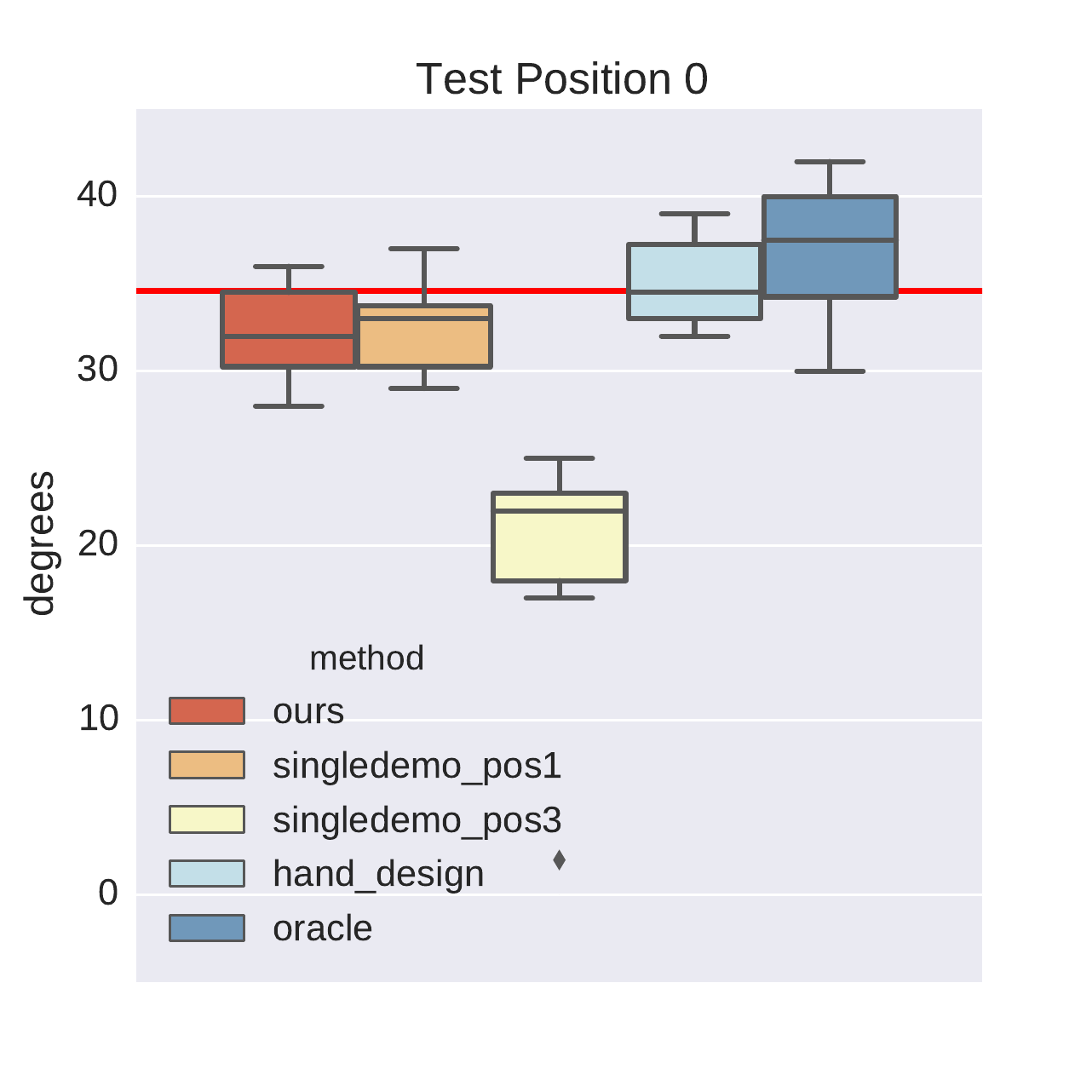}
    \hfill
    \includegraphics[width=.245\textwidth]{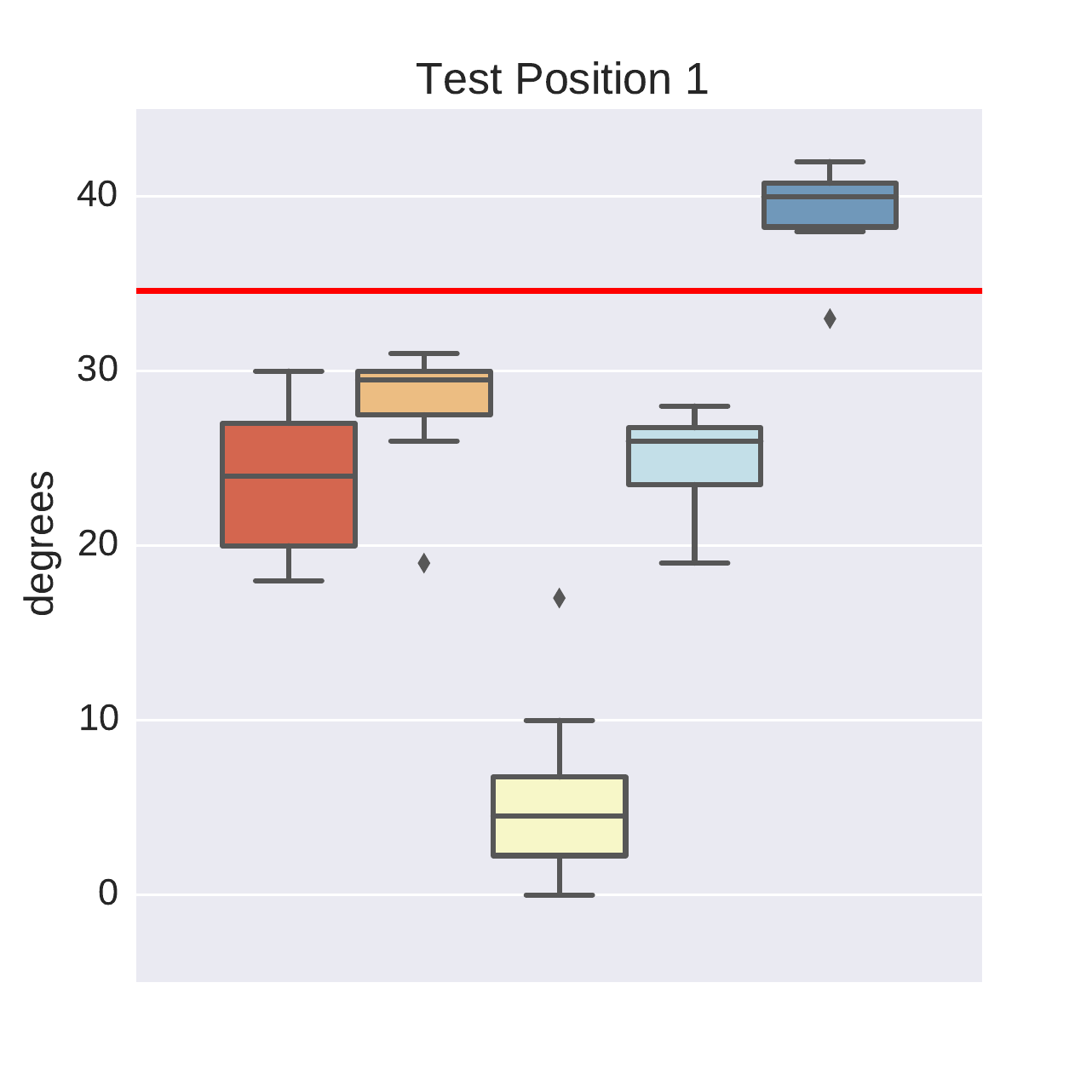}
    \hfill
    \includegraphics[width=.245\textwidth]{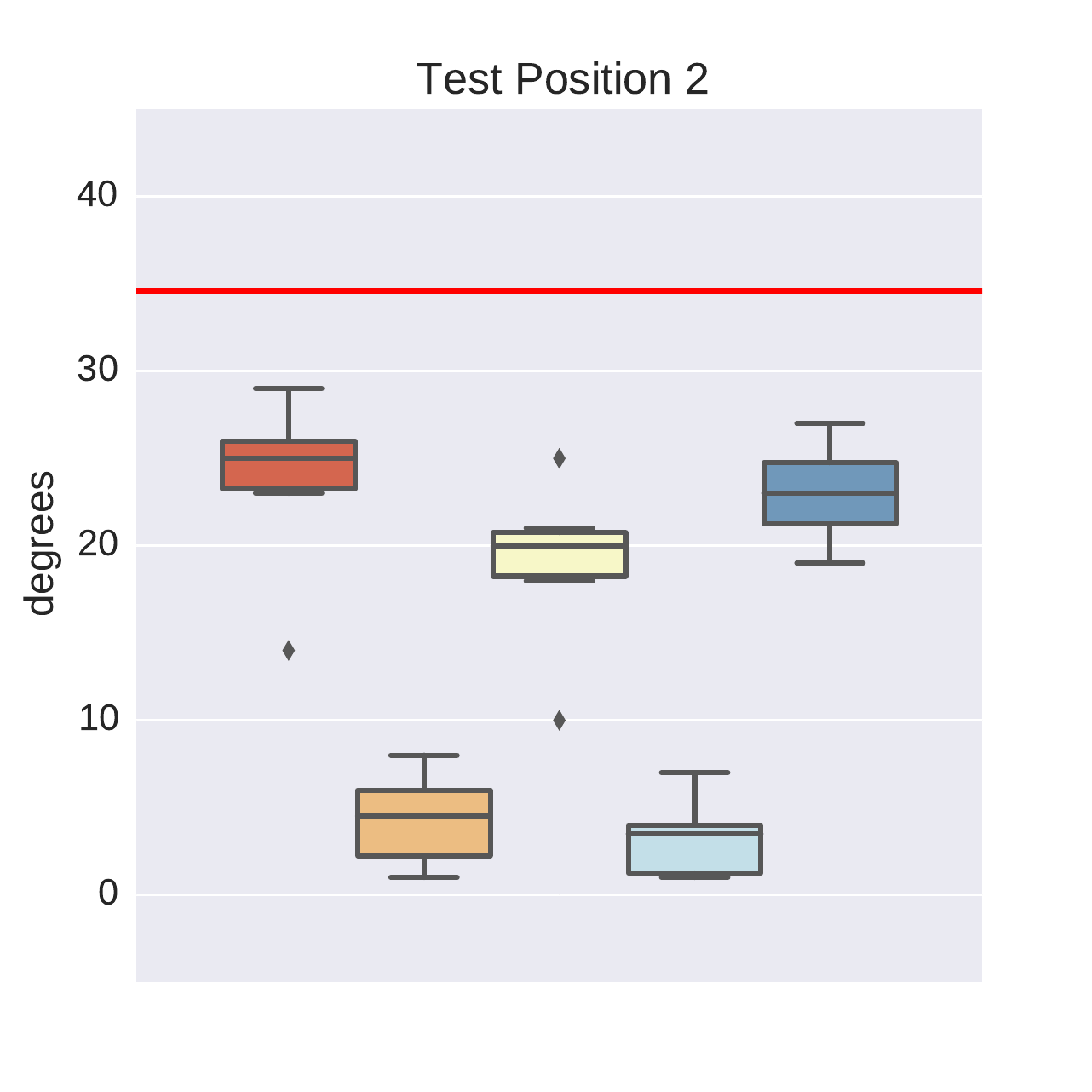}
    \hfill
    \includegraphics[width=.245\textwidth]{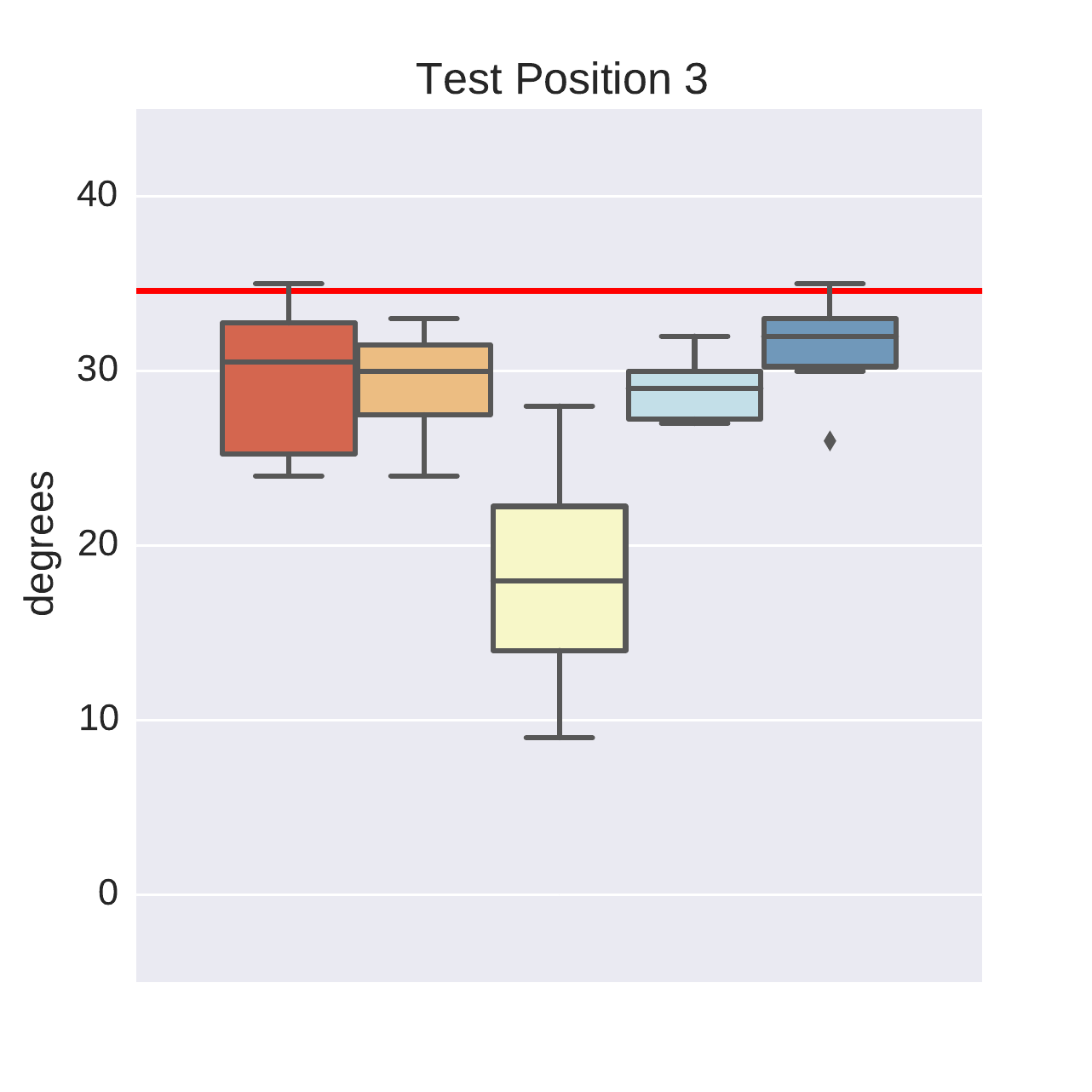}
    \vspace{-1cm}
    \caption{Comparison of different policies for the valve task: the red line indicates the demonstrated rotation of the valve by $\approx 35 \deg$. On average our method learns the most general feedback strategy. The boxes in the box plot for each test position are our method, single demo baseline 1, single demo baseline 2, hand-designed baseline and oracle plotted from left to right. Although the baselines do well in some positions, the only methods which do consistently well across all positions are our method and the oracle.}
    \label{fig:valvetask_comparison}
\end{figure*}

\begin{figure*}
\begin{center}
\raisebox{-0.45\height}{\includegraphics[height=1.5cm]{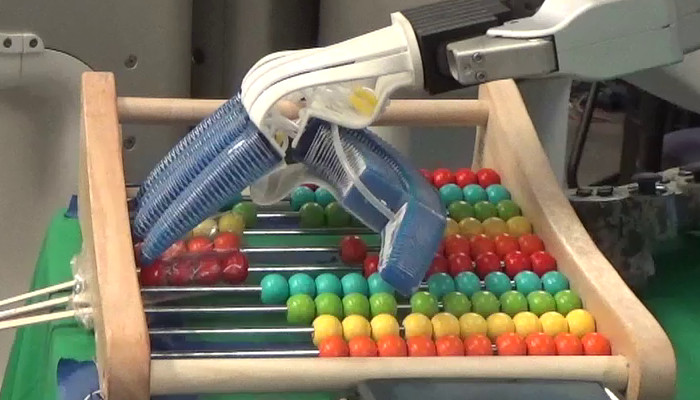}}
\resizebox{.75\textwidth}{!}{%
 \begin{tabular}{||l c c c c c c c||} 
 \hline
 Bead & Target & \textbf{Ours} & SingleDemo1 & SingleDemo2 & Oracle & HandDesign1 & HandDesign2 \\ [0.5ex] 
 \hline\hline
 1 & 8.4 & \textbf{7.49 $\pm$ 0.47} & 7.02 $\pm$ 0.50 & 6.33 $\pm$ 2.15 & 7.66 $\pm$ 0.23 & 8.38 $\pm$ 0.04 & 0 $\pm$ 0\\ 
 \hline
 2 & 0 & \textbf{0.14 $\pm$ 0.18} & 0.60 $\pm$ 0.69 & 7.08 $\pm$ 1.04 & 0.27 $\pm$ 0.42 & 0 $\pm$ 0 & 6.5 $\pm$ 0\\
 \hline
 3 & 0 & \textbf{0.89 $\pm$ 1.00} & 0.28 $\pm$ 0.18 & 1.23 $\pm$ 2.20 & 1.08 $\pm$ 0.72 & 0 $\pm$ 0 & 8.43 $\pm$ 0.29 \\
 \hline
 \end{tabular}}
 
\begin{center}
\raisebox{-0.45\height}{\includegraphics[height=1.5cm]{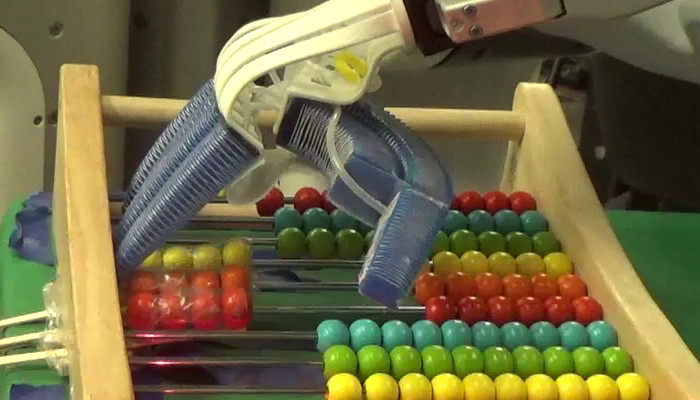}}
\resizebox{.75\textwidth}{!}{%
 \begin{tabular}{||l c c c c c c c||} 
 \hline
 Bead & Target & \textbf{Ours} & SingleDemo1 & SingleDemo2 & Oracle & HandDesign1 & HandDesign2 \\ [0.5ex] 
 \hline\hline
 1 & 8.4 & \textbf{7.95 $\pm$ 0.19} & 1.04 $\pm$ 2.15 & 7.27 $\pm$ 0.65 & 7.52 $\pm$ 0.66 & 0.00 $\pm$ 0.00 & 8.38 $\pm$ 0.08 \\ 
 \hline 
 2 & 0 & \textbf{0.10 $\pm$ 0.10} & 0.85 $\pm$ 1.21 & 0.19 $\pm$ 0.14 & 0.09 $\pm$ 0.11 & 0.00 $\pm$ 0.00 & 8.40 $\pm$ 0.00 \\ 
 \hline 
 3 & 0 & \textbf{0.00 $\pm$ 0.00} & 0.00 $\pm$ 0.00 & 0.00 $\pm$ 0.00 & 0.00 $\pm$ 0.00 & 0.00 $\pm$ 0.00 & 0.00 $\pm$ 0.00 \\ 
 \hline
 \end{tabular}} 
\end{center}

\begin{center}
\raisebox{-0.45\height}{\includegraphics[height=1.5cm]{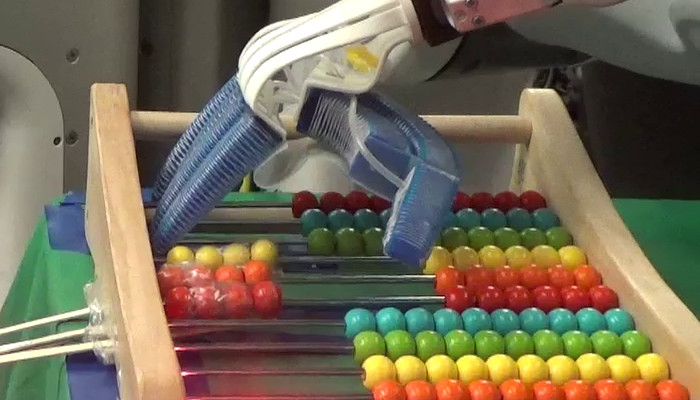}}
\resizebox{.75\textwidth}{!}{%
 \begin{tabular}{||l c c c c c c c||} 
 \hline
 Bead & Target & \textbf{Ours} & SingleDemo1 & SingleDemo2 & Oracle & HandDesign1 & HandDesign2 \\ [0.5ex] 
 \hline\hline
 1 & 8.4 & \textbf{7.21 $\pm$ 0.69} & 2.47 $\pm$ 2.22 & 3.39 $\pm$ 1.98 & 7.74 $\pm$ 0.23 & 0.00 $\pm$ 0.00 & 8.38 $\pm$ 0.05 \\ 
 \hline 
 2 & 0 & \textbf{0.00 $\pm$ 0.00} & 0.00 $\pm$ 0.00 & 0.00 $\pm$ 0.00 & 0.00 $\pm$ 0.00 & 0.00 $\pm$ 0.00 & 0.00 $\pm$ 0.00 \\ 
 \hline 
 3 & 0 & \textbf{0.00 $\pm$ 0.00} & 0.00 $\pm$ 0.00 & 0.00 $\pm$ 0.00 & 0.00 $\pm$ 0.00 & 0.00 $\pm$ 0.00 & 0.00 $\pm$ 0.00 \\ 
 \hline 
 \end{tabular}} 
\end{center}

\end{center}
    \caption{Comparison of the distance moved by the various beads in cm using different policies for the abacus task, at 3 different positions, namely Positions 1, 2 and 3 going downwards. The target column in each table indicates the demonstrated movement of the three beads, and the other columns indicate the mean and standard deviation of other methods. On average our method learns the most general feedback strategy besides the oracle}
    \label{fig:abacustask_comparison}
    \vspace{\gapbetweencaptionandtext}
\end{figure*}

\subsubsection{Results and Discussion}
During evaluation, the policy learned by each method was sampled ten times at four positions of the hand relative to the valve.
The results in Fig.~\ref{fig:valvetask_comparison} show that our method generates the most robust policy compared with the baselines, which each fail to turn the valve significantly in at least one position.
Our method does nearly as well as the oracle, for which demonstrations are assigned to controllers manually.
While learning the correspondence weights and the individual controller policies, our method determines which of the demonstration it can actually perform from its initial positions, and disregards distant unachievable demonstrations. 

Our method is able to learn distinctly different behavior at various test positions. At position~1, the policy pushes the lever using its last two fingers, with support given by the thumb. At position~2, the policy uses the thumb to rotate the valve by pushing the lever as the fingers are blocked. At position~3, our policy extends the thumb out of the way and pushes strongly with the index finger to rotate the valve.
Simple open loop hand-designed strategies and the baselines learned from a single demonstration fail to learn this distinctly different behavior needed to generalize to different positions along the valve lever. By learning that different joint angles of the arm require different behaviors to be performed, our method is able to perform the task in various positions.


\subsection{Pushing the beads of an abacus}
\subsubsection{Experimental setup}
The RBO Hand~2 is required to push particular beads on an abacus while leaving other beads stationary. This task is challenging due to the precise individual finger motions needed to move only the desired beads. The hand is mounted on a stationary PR2 arm (Fig.~\ref{fig:experimentalsetups}), while the abacus is moved to several positions. The beads of relevance here are the central yellow, orange and red ones. Markers were attached to each of the three beads to capture their motion. As the position of the abacus with respect to the hand changes, different fingers need to be used.
During demonstrations a human pushed only the yellow beads along their spindle at each of the three positions shown in Fig.~\ref{fig:abacustask_comparison}.
\begin{figure*}[h]
    \centering
    \includegraphics[width=.16\textwidth]{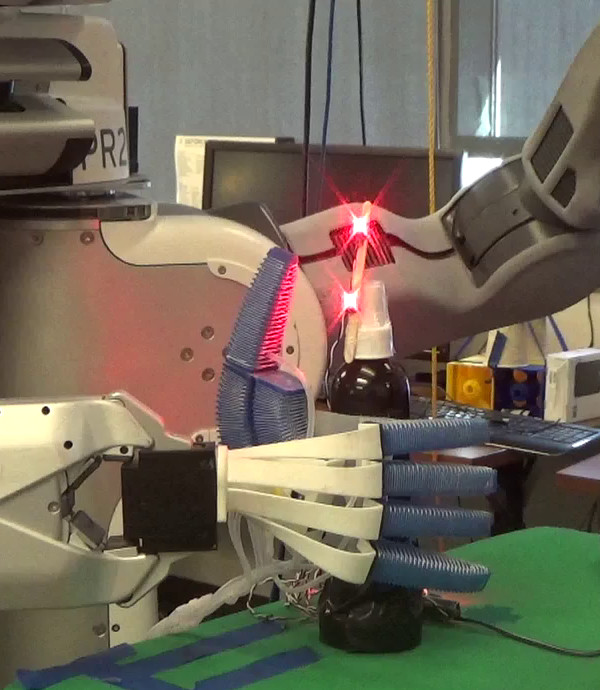}
    \hfill
    \includegraphics[width=.16\textwidth]{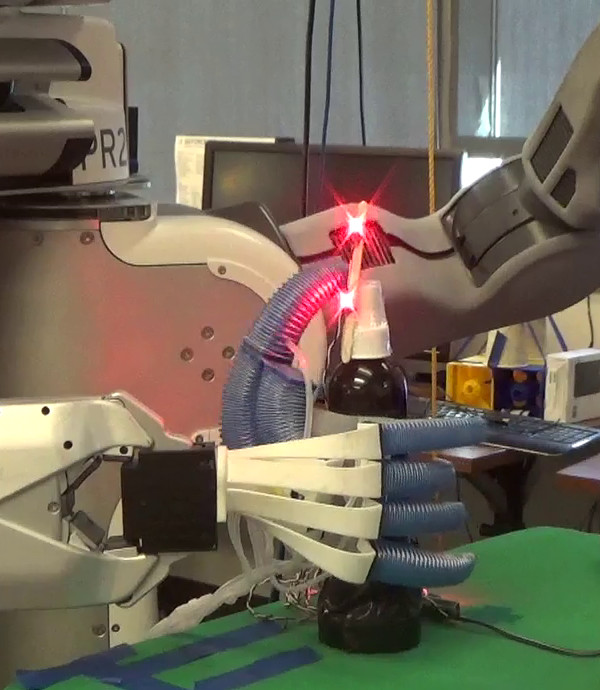}
    \hfill
    \includegraphics[width=.16\textwidth]{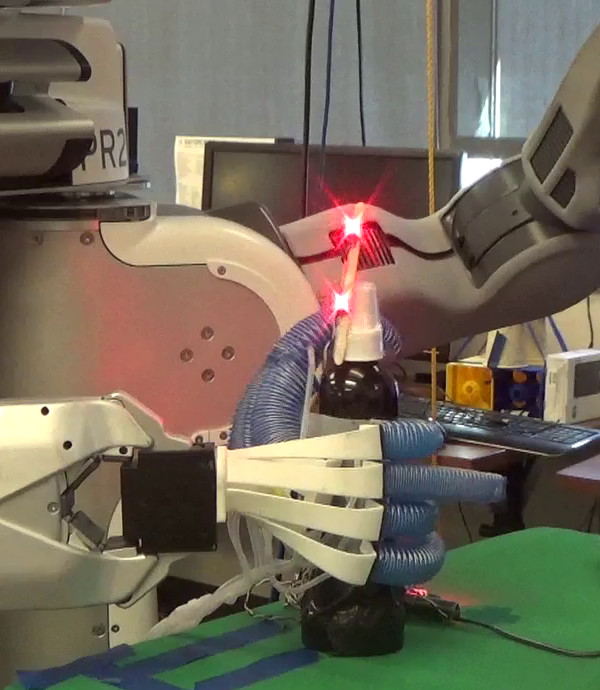}
    \hfill
    \includegraphics[width=.16\textwidth]{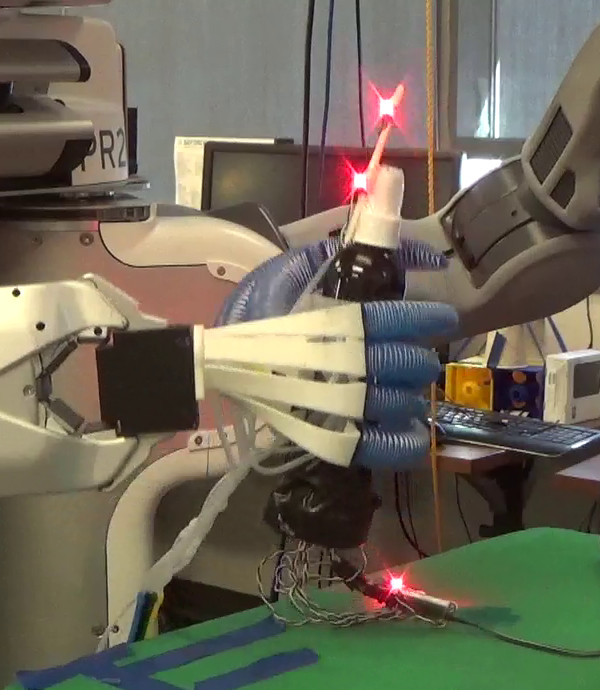}
    \hfill
    \includegraphics[width=.16\textwidth]{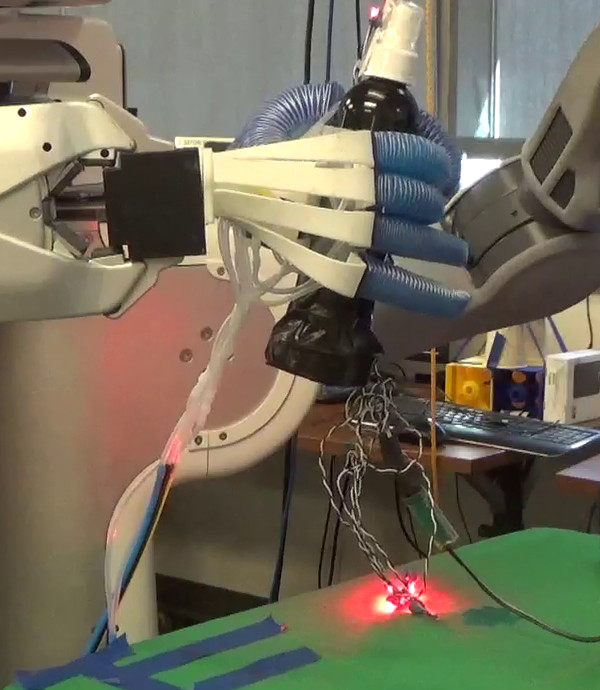}
    \caption{Execution of a learned policy to grasp a bottle. Our method learns to grasp the bottle tightly and performs as well as the hand designed baseline.}
    \label{fig:bottlegrasping}
    \vspace{\gapbetweencaptionandtext}
\end{figure*}
\subsubsection{Results and Discussion}
We evaluated ten samples of each policy at each of the three positions and recorded the distances that each bead moved. The results are shown in Fig.~\ref{fig:abacustask_comparison}. Our method moves the bead closer to the target position than the single demonstration and hand designed baselines for all the test positions. Only the oracle policy produces equally good performance. By interleaving selection of the right demonstration to imitate, with optimal control and supervised learning, our algorithm is able to learn a policy which uses discretely different fingers depending on the positions of the abacus relative to the hand.
On the other hand, the hand-designed baselines being open loop can never learn different behaviors for different fingers. The controller trained at a single position fails because it has no notion of generalization.


\subsection{Grasping a bottle}

\subsubsection{Experimental setup}
This task involves using the soft hand mounted on a moving arm, to grasp a deodorant bottle placed on a table. The arm has a scripted motion of moving up after 8 seconds, and we use reinforcement learning to learn the behavior of the fingers to go with this arm motion. The objective of the task is to grasp the bottle before the arm starts moving and keep it grasped until the end of the episode at the final arm location.

As grasping tasks for several objects largely succeed in open loop, our aim is to demonstrate that we can match the performance of a hand-designed baseline with a controller learned from a human demonstration through optimal control. This experiment is challenging for reinforcement learning algorithms due to the delayed nature of the reward signal in grasping. 

We provide a demonstration of the bottle being lifted by a human, and use it to define the cost function for trajectory optimization as the $l_2$ distance of trajectory samples from the provided demonstration. We also apply a Gaussian filter to the noise generated in the controllers to be more temporally coherent, allowing tight grasping.
The resulting learned control policy is then tested on 10 sample trajectories in order to evaluate whether a successful grasp has occurred where the objected is lifted and kept at the maximum arm height. 

\subsubsection{Results and Discussion}

We find that on the grasping task, the control policy learned through optimal control does just as well as a hand-designed policy on ten samples of grasping the bottle. Both the hand-designed policy and the learned policy were able to grasp the bottle for all 10 test samples.
This indicates that the learning has comparable results to a hand-designed baseline, despite not having prior information besides a human-provided demonstration.


\subsection{Limitations}
Although the LfD algorithm shows good performance on several tasks using the RBO Hand~2, there are many directions for future work. 
Instead of using a motion capture system, we hope to
use better computer vision techniques such as deep convolutional nets to track trajectories of relevant feature points in future work.
Extending the neural network policy to learn policies dependent on just the pressure sensors in the fingers and/or additional tactile sensors, would be an exciting future direction.

\section{Conclusions}

We presented an algorithm for learning dexterous manipulation skills with a soft hand from object-centric demonstrations. Unlike standard LfD methods, our approach only requires the human expert to demonstrate the desired behaviors with their own hand. Our method automatically determines the most relevant demonstrations to track, using reinforcement learning to optimize a collection of controllers together with controller to demonstration correspondences. To generalize the demonstrations to new initial conditions, we utilize the GPS framework to train nonlinear neural network policies that combine the capabilities of all of the controllers. We evaluate our method on the RBO Hand~2, and show that it is capable of learning a variety of dexterous manipulation skills, including valve turning, moving beads on an abacus, and grasping.

\addtolength{\textheight}{-12cm}   

\bibliographystyle{IEEEtran}
\bibliography{references}

\begin{thebibliography}{10}
\providecommand{\url}[1]{#1}
\csname url@samestyle\endcsname
\providecommand{\newblock}{\relax}
\providecommand{\bibinfo}[2]{#2}
\providecommand{\BIBentrySTDinterwordspacing}{\spaceskip=0pt\relax}
\providecommand{\BIBentryALTinterwordstretchfactor}{4}
\providecommand{\BIBentryALTinterwordspacing}{\spaceskip=\fontdimen2\font plus
\BIBentryALTinterwordstretchfactor\fontdimen3\font minus
  \fontdimen4\font\relax}
\providecommand{\BIBforeignlanguage}[2]{{%
\expandafter\ifx\csname l@#1\endcsname\relax
\typeout{** WARNING: IEEEtran.bst: No hyphenation pattern has been}%
\typeout{** loaded for the language `#1'. Using the pattern for}%
\typeout{** the default language instead.}%
\else
\language=\csname l@#1\endcsname
\fi
#2}}
\providecommand{\BIBdecl}{\relax}
\BIBdecl

\bibitem{mouri2002anthropomorphic}
T.~Mouri, H.~Kawasaki, K.~Yoshikawa, J.~Takai, and S.~Ito, ``Anthropomorphic
  robot hand: Gifu hand iii,'' in \emph{Proc. Int. Conf. ICCAS}, 2002.

\bibitem{shadowhand}
``Shadow robot company. the shadow dexterous hand.''
  http://www.shadowrobot.com/products/dexterous-hand/.

\bibitem{Deimel16-IJRR}
R.~Deimel and O.~Brock, ``A novel type of compliant and underactuated robotic
  hand for dexterous grasping,'' \emph{The International Journal of Robotics
  Research}, vol.~35, no. 1-3, pp. 161--185, March 2016.

\bibitem{levine2014learning}
S.~Levine and P.~Abbeel, ``Learning neural network policies with guided policy
  search under unknown dynamics,'' in \emph{NIPS}, 2014.

\bibitem{han1998dextrous}
L.~Han and J.~C. Trinkle, ``Dextrous manipulation by rolling and finger
  gaiting,'' in \emph{ICRA}, vol.~1.\hskip 1em plus 0.5em minus 0.4em\relax
  IEEE, 1998.

\bibitem{cherif1999planning}
M.~Cherif and K.~K. Gupta, ``Planning quasi-static fingertip manipulations for
  reconfiguring objects,'' \emph{Robotics and Automation, IEEE Transactions
  on}, vol.~15, no.~5, pp. 837--848, 1999.

\bibitem{bai2014dexterous}
Y.~Bai and C.~K. Liu, ``Dexterous manipulation using both palm and fingers,''
  in \emph{ICRA 2014}.\hskip 1em plus 0.5em minus 0.4em\relax IEEE, 2014.

\bibitem{mordatch2012contact}
I.~Mordatch, Z.~Popovi{\'c}, and E.~Todorov, ``Contact-invariant optimization
  for hand manipulation,'' in \emph{Proc. of the ACM SIGGRAPH}, 2012.

\bibitem{polygerinos2015fem}
P.~Polygerinos, Z.~Wang, J.~T.~B. Overvelde, K.~C. Galloway, R.~J. Wood,
  K.~Bertoldi, and C.~J. Walsh, ``Modeling of soft fiber-reinforced bending
  actuators,'' \emph{IEEE T-RO}, vol.~31, no.~3, pp. 778--789, June 2015.

\bibitem{ijspeert2002learning}
A.~J. Ijspeert, J.~Nakanishi, and S.~Schaal, ``Learning attractor landscapes
  for learning motor primitives,'' Tech. Rep., 2002.

\bibitem{peters2010relative}
J.~Peters, K.~M{\"u}lling, and Y.~Altun, ``Relative entropy policy search.'' in
  \emph{AAAI}.\hskip 1em plus 0.5em minus 0.4em\relax Atlanta, 2010.

\bibitem{theodorouAISTATS2010}
B.~J. S.~S. Theodorou, E.~A., ``Learning policy improvements with path
  integrals,'' in \emph{AISTATS 2010}, 2010.

\bibitem{mulling2013learning}
K.~M{\"u}lling, J.~Kober, O.~Kroemer, and J.~Peters, ``Learning to select and
  generalize striking movements in robot table tennis,'' \emph{IJRR}, vol.~32,
  no.~3, pp. 263--279, 2013.

\bibitem{kroemer2015towards}
O.~Kroemer, C.~Daniel, G.~Neumann, H.~van Hoof, and J.~Peters, ``Towards
  learning hierarchical skills for multi-phase manipulation tasks,'' in
  \emph{ICRA}.\hskip 1em plus 0.5em minus 0.4em\relax IEEE, 2015.

\bibitem{kalakrishnan2011learning}
M.~Kalakrishnan, L.~Righetti, P.~Pastor, and S.~Schaal, ``Learning force
  control policies for compliant manipulation,'' in \emph{IROS}.\hskip 1em plus
  0.5em minus 0.4em\relax IEEE, 2011.

\bibitem{lampe2013acquiring}
T.~Lampe and M.~Riedmiller, ``Acquiring visual servoing reaching and grasping
  skills using neural reinforcement learning,'' in \emph{IJCNN}, 2013.

\bibitem{levine2015learning}
S.~Levine, N.~Wagener, and P.~Abbeel, ``Learning contact-rich manipulation
  skills with guided policy search,'' in \emph{ICRA}, 2015.

\bibitem{levine2015end}
S.~Levine, C.~Finn, T.~Darrell, and P.~Abbeel, ``End-to-end training of deep
  visuomotor policies,'' \emph{arXiv preprint arXiv:1504.00702}, 2015.

\bibitem{vanHoof2015learning}
H.~van Hoof, T.~Hermans, G.~Neumann, and J.~Peters, ``Learning robot in-hand
  manipulation with tactile features,'' in \emph{Humanoid Robots
  (Humanoids)}.\hskip 1em plus 0.5em minus 0.4em\relax IEEE, 2015.

\bibitem{kumaroptimal}
V.~Kumar, E.~Todorov, and S.~Levine, ``Optimal control with learned local
  models: Application to dexterous manipulation.''

\bibitem{calinon2007incremental}
S.~Calinon and A.~Billard, ``Incremental learning of gestures by imitation in a
  humanoid robot,'' in \emph{Proceedings of the ACM/IEEE international
  conference on Human-robot interaction}.\hskip 1em plus 0.5em minus
  0.4em\relax ACM, 2007.

\bibitem{asfour2008imitation}
T.~Asfour, P.~Azad, F.~Gyarfas, and R.~Dillmann, ``Imitation learning of
  dual-arm manipulation tasks in humanoid robots,'' \emph{International Journal
  of Humanoid Robotics}, vol.~5, no.~02, pp. 183--202, 2008.

\bibitem{sauser2012iterative}
E.~L. Sauser, B.~D. Argall, G.~Metta, and A.~G. Billard, ``Iterative learning
  of grasp adaptation through human corrections,'' \emph{Robotics and
  Autonomous Systems}, vol.~60, no.~1, pp. 55--71, 2012.

\bibitem{prieur2012modeling}
U.~Prieur, V.~Perdereau, and A.~Bernardino, ``Modeling and planning high-level
  in-hand manipulation actions from human knowledge and active learning from
  demonstration,'' in \emph{IROS}.\hskip 1em plus 0.5em minus 0.4em\relax IEEE,
  2012.

\bibitem{guenter2007reinforcement}
F.~Guenter, M.~Hersch, S.~Calinon, and A.~Billard, ``Reinforcement learning for
  imitating constrained reaching movements,'' \emph{Advanced Robotics},
  vol.~21, no.~13, pp. 1521--1544, 2007.

\bibitem{hershey2007kldmixtures}
J.~Hershey and P.~Olsen, ``Approximating the {K}ullback {L}eibler divergence
  between {G}aussian mixture models,'' in \emph{ICASSP}, vol.~4, April 2007,
  pp. IV--317--IV--320.

\bibitem{opac-b1078358}
D.~H. Jacobson and D.~Q. Mayne, \emph{Differential dynamic programming}.\hskip
  1em plus 0.5em minus 0.4em\relax New York: Elsevier, 1970.

\bibitem{levine2013guided}
S.~Levine and V.~Koltun, ``Guided policy search,'' in \emph{Proceedings of The
  30th International Conference on Machine Learning}, 2013.

\bibitem{li2004iterative}
W.~Li and E.~Todorov, ``Iterative linear quadratic regulator design for
  nonlinear biological movement systems.'' in \emph{ICINCO (1)}, 2004.

\bibitem{kapandji_cotation_1986}
I.~A. Kapandji, ``Cotation clinique de l’opposition et de la
  contre-opposition du pouce,'' \emph{Annales de Chirurgie de la Main}, vol.~5,
  no.~1, pp. 68--73, 1986.

\end{thebibliography}

\end{document}